\documentclass{article}

\usepackage{arxiv}

\usepackage[utf8]{inputenc} 
\usepackage[T1]{fontenc}    
\usepackage{hyperref}       
\usepackage{url}            
\usepackage{booktabs}       
\usepackage{amsfonts}       
\usepackage{nicefrac}       
\usepackage{microtype}      
\usepackage{lipsum}		
\usepackage{graphicx}
\usepackage{natbib}
\usepackage{doi}
\usepackage{amsmath, amssymb}
\usepackage{amsfonts}
\usepackage{multirow}
\usepackage{balance} 
\usepackage{array}

\title{On Evaluating Loss Functions for Stock Ranking: \\An Empirical Analysis With Transformer Model}

\author{ \href{https://orcid.org/0009-0006-6071-1607}{\hspace{1mm}Jan Kwiatkowski} \\
	Institute of Computer Science, Warsaw University of Technology\\
	\texttt{jan.kwiatkowski.stud@pw.edu.pl} \\
	\And
	\href{https://orcid.org/0000-0003-4534-8652}{\hspace{1mm}Jarosław A. Chudziak} \\
	Institute of Computer Science, Warsaw University of Technology \\
    \texttt{jaroslaw.chudziak@pw.edu.pl} \\
}

\begin{document}
\maketitle

\begin{abstract}
    Quantitative trading strategies rely on accurately ranking stocks to identify profitable investments. Effective portfolio management requires models that can reliably order future stock returns. Transformer models are promising for understanding financial time series, but how different training loss functions affect their ability to rank stocks well is not yet fully understood.
Financial markets are challenging due to their changing nature and complex relationships between stocks. Standard loss functions, which aim for simple prediction accuracy, often aren't enough. They don't directly teach models to learn the correct order of stock returns. While many advanced ranking losses exist from fields such as information retrieval, there hasn't been a thorough comparison to see how well they work for ranking financial returns, especially when used with modern Transformer models for stock selection.
This paper addresses this gap by systematically evaluating a diverse set of advanced loss functions including pointwise, pairwise, listwise for daily stock return forecasting to facilitate rank-based portfolio selection on S$\&$P 500 data. We focus on assessing how each loss function influences the model's ability to discern profitable relative orderings among assets.
Our research contributes a comprehensive benchmark revealing how different loss functions impact a model's ability to learn cross-sectional and temporal patterns crucial for portfolio selection, thereby offering practical guidance for optimizing ranking-based trading strategies.
\end{abstract}

%
%



\maketitle

\section{Introduction}\label{sec:intro}
Quantitative trading strategies now use advanced machine learning models to find profitable investment opportunities in financial markets \citep{quantitative_investment_survey, quantformer}. These models help traders make better decisions by analyzing complex market data and predicting which stocks will perform well. 
One key task in portfolio management \citep{bankachudziak_deltaHeghe} is ranking stocks based on how well they are expected to perform in the future \citep{Stock_Ranking_with_MTL}. Getting this ranking right is crucial because it directly affects how money is allocated across different stocks. This allocation decision determines both the profits and risks of the final portfolio \citep{perrin2019machinelearningoptimizationalgorithms}.
Traditional statistical models, such as ARIMA, have been used for financial time series analysis and forecasting for many years \citep{siaminamini2018forecastingeconomicsfinancialtime}. However, deep learning models, especially Transformer architectures \citep{transformer}, offer new possibilities. They excel in capturing complex patterns and long-range dependencies in sequential data \citep{deep_time_series_models_benchmark}. This makes them well suited for analyzing financial time series data, where it is important to understand trends and relationships between different assets \citep{tft}. 

Financial markets are difficult to predict because they are noisy, relationships between assets change over time, and the patterns that drive stock prices are subtle and constantly evolving \citep{GVK483463442, 10.5555/3217448,Diffusion_model_stock_prediction}. Therefore, to successfully apply Transformers to financial forecasting, both the structure of the model and the way the model learns from the data must be carefully designed.

The way a model learns is controlled by its loss function \citep{Comprehensive_Survey_of_Los_FunctionsinML}. This function defines the prediction error that the model tries to minimize during training. For stock ranking tasks, correctly identifying which stocks will outperform others is more important than predicting exact future returns with perfect accuracy \citep{alsp_tf}. A model that makes small prediction errors but gets the ranking wrong may be less useful than one that makes larger errors but correctly identifies the best investment opportunities.
While advanced ranking-oriented loss functions categorized as pointwise, pairwise, and listwise have been extensively studied and successfully applied in fields like information retrieval \citep{10.1561/1500000016} and more recently in tasks like Neural Architecture Search (NAS) performance prediction \citep{ji2024evaluating}, their evaluation and adaptation for financial stock ranking using Transformer models remain underexplored. 

To address this gap, we conduct a comprehensive study of different loss functions for training Transformer based architecture (PortfolioMASTER, inspired by \citet{master}) for daily stock return ranking on S$\&$P 500 data. Our goal is to understand how different loss functions affect the model's ability to learn patterns in stock data and generate profitable rankings.
This investigation contributes a comprehensive benchmark of pointwise, pairwise, listwise, and weighted ranking loss functions for Transformer based stock selection. The results offer insights into how different loss functions influence the model's ability to leverage cross-sectional relationships between stocks and temporal patterns for effective portfolio selection.
By clarifying the relationship between loss function choice and portfolio performance, this study provides clearer direction for developing more effective deep learning models for quantitative finance.

\section{Related work}\label{sec:related_work}
This research intersects with several key areas: deep learning for financial forecasting, the application of Transformer models to time series, Learning-to-Rank (LTR) methodologies, and the role of specific loss functions in ranking tasks.

The use of deep learning for financial market prediction has seen significant growth. Initial explorations involved models such as Recurrent Neural Networks (RNNs), Long Short-Term Memory networks (LSTMs) \citep{FISCHER2018654}, and Convolutional Neural Networks (CNNs) \citep{10.5555/2832415.2832572, CNN_BiLSTM_AM} for tasks like stock price prediction \citep{ml_approaches_in_stock_price_prediction}. More recently, Transformer architectures \citep{transformer}, originally designed for natural language processing, have been successfully adapted for general time series forecasting \citep{zhou2021informerefficienttransformerlong, stockmixer, iTransformer, deep_time_series_models_benchmark}. Consequently, a growing body of work is now exploring the application of these Transformer models specifically within the financial domain \citep{banka2025informer, szydłowski2024hidformertransformerstyleneuralnetwork, hedge_itransformer, 10849476,Galformer, Lamformer2025}.

Learning-to-Rank (LTR), a field predominantly developed for information retrieval (IR) tasks such as document ranking for search engines \citep{10.1561/1500000016, Li2011LTRSurvey}, offers three main strategies: pointwise, pairwise, and listwise. The \textit{pointwise approach} treats ranking as a regression or classification problem on individual items, predicting a score for each, which then dictates the final order \citep{Cossock2006SubsetRanking}. In contrast, the \textit{pairwise approach} focuses on the relative order of item pairs. The model learns to predict whether item $i$ should be ranked higher than item $j$. Loss functions like RankNet \citep{Burges2005RankNet}, Margin Ranking Loss \citep{Herbrich2000SVMRegression}, and Bayesian Personalized Ranking (BPR) \citep{Rendle2009BPR} fall into this category. Finally, the \textit{listwise approach} considers the entire list of items as a single instance for training and directly optimizes a measure of the quality of the entire ranked list \citep{Stock_Ranking_Prediction_Using_ListWise}, with examples such as ListNet \citep{Cao2007ListNet} and ListMLE \citep{Xia2008ListMLE}. 

The critical role of the loss function in ranking tasks extends beyond IR. For instance, in Neural Architecture Search (NAS), performance predictors are used to estimate the effectiveness of candidate neural architectures. Recent studies have shown \citet{ji2024evaluating} that replacing standard Mean Squared Error (MSE) with pairwise or listwise ranking losses significantly improves the predictors ability to identify top performing architectures. \citet{ji2024evaluating} benchmarked 11 ranking losses for NAS, concluding that the choice of loss function affects predictor quality and that rank metrics are strong indicators for discovering optimal architectures. 

Furthermore, specialized loss functions like WARP loss \citep{10.5555/2283696.2283856} and LambdaLoss \citep{10.1145/3269206.3271784} have been developed in IR to improve how models rank results, showing that choosing the right loss function for the task can improve performance. Our work aims to bridge the gap by systematically evaluating a range of these ranking loss functions for Transformer based stock selection.

\section{Methodology}\label{sec:problem_definition}
In this chapter, we explain the methodological approach for our study. We begin by defining the stock ranking task, then describe the Transformer architecture used for prediction, and finally, we present the different loss functions we tested, which are the core subject of this analysis.

\begin{table*}[htbp!]
\centering
\caption{Overview of Investigated Loss Functions and their Key Components. For Combined Losses, $L_{\text{PairwiseComp.}}$ is shown, which is used in Eq.~\ref{eq:combined_loss_defined_here}. Normalization over pairs (e.g., $1/|P_{\text{valid}}|$) is applied. Let $s_{ij} = \text{sign}(y_i - y_j)$.}
\label{tab:loss_function_formulas_simplified}
\resizebox{\textwidth}{!}{
    \begin{tabular}{@{}l l l l@{}}
    \toprule
    \textbf{Loss Name} & \textbf{Category} & \multicolumn{1}{c}{\textbf{Core Component Formula}} & \textbf{Tunable Params} \\
    \midrule
    MSE (Baseline) & Pointwise & $L_{\text{MSE}}(\hat{\mathbf{y}}, \mathbf{y})$ \text{ (see Eq.~\ref{eq:mse_defined_here})} & \text{None} \\
    \midrule
    \multicolumn{4}{l}{\textit{Combined Losses (Pairwise Component $L_{\text{PairwiseComp.}}$ shown)}} \\
    \midrule
    Hinge & Pointwise + Pairwise & $\sum_{(i,j) \in P_{\text{valid}}} \max(0, m - s_{ij}(\hat{y}_i - \hat{y}_j))$ & $\lambda$, \text{margin }$m$ \\
    Margin & Pointwise + Pairwise & $\sum_{(i,j) \in P_{\text{valid}}} \max(0, -s_{ij}(\hat{y}_i - \hat{y}_j) + m)$ & $\lambda$, \text{margin }$m$ \\
    BPR & Pointwise + Pairwise & $\sum_{\{(i,j) \mid y_i > y_j\}} \log(1 + \exp(-(\hat{y}_i - \hat{y}_j)))$ & $\lambda$ \\
    RankNet & Pointwise + Pairwise & $\sum_{(i,j) \in P_{\text{valid}}} \log(1 + \exp(-\alpha \cdot s_{ij} (\hat{y}_i - \hat{y}_j)))$ & $\lambda$, \text{scale }$\alpha$ \\
    WHR1/WHR2 & Pointwise + Pairwise (Weighted Hinge) & $\sum_{(i,j) \in P_{\text{valid}}} w_i w_j \max(0, m - s_{ij} (\hat{y}_i - \hat{y}_j))$ \newline \quad \text{where }$w_k^{\text{WHR1/2}}$ \text{ based on rank} & $\lambda$, $\text{margin}$ $m$ \\
    & & \multicolumn{1}{l}{\quad \text{where }$w_k^{\text{WHR1/2}}$ \text{ based on rank}} & \\ 
    \midrule
    \multicolumn{4}{l}{\textit{Listwise Loss}} \\
    \midrule
    ListNet & Listwise & $L_{\text{ListNet}}(\hat{\mathbf{y}}, \mathbf{y}; \tau)$ \text{ (see Eq.~\ref{eq:listnet_final_loss})} & \text{Temperature }$\tau$ \\
    \bottomrule
    \end{tabular}
}
\end{table*}

\subsection{Problem Definition}
\label{sec:problem_definition}
Quantitative portfolio selection aims to optimize asset allocation by predicting future stock returns and then ranking stocks from best to worst based on their expected performance. The predictive model is trained to capture temporal and cross-sectional stock behaviors by minimizing a specific loss function, $L_j$. The choice of $L_j$ fundamentally shapes how the model interprets error and what relationships it prioritizes. However, while many advanced ranking losses exist, their effectiveness in training modern Transformer models for this specific task remains underexplored. This paper addresses this gap by investigating the impact of various ranking loss functions on the performance of a Transformer based model for rank based stock selection.

Our core task is stock return forecasting formulated as a ranking problem~\cite{alsp_tf, ci-sthpan}. We consider a universe of $N$ assets. For each trading day $t$, the input to our model is $X_t \in \mathbb{R}^{T \times N \times F}$, where $T$ is the lookback window, $N$ the number of assets, and $F$ the number of input features. The model, PortfolioMASTER, learns $f_{\theta, L_j}: \mathbb{R}^{T \times N \times F} \rightarrow \mathbb{R}^{N}$, mapping $X_t$ to predicted returns $\hat{R}_{t+1} = [\hat{r}_{1}^{t+1}, \dots, \hat{r}_{N}^{t+1}]$. The parameters $\theta$ are optimized by minimizing $L_j$ from a set of investigated losses (Section~\ref{sec:loss_functions}). The ground truth target $r_{i}^{t+1}$ is the next day return for stock $s_i$: $r_{i}^{t+1} = (p_{i}^{t+1} - p_{i}^{t}) / p_{i}^{t}$, where $p_{i}^{t}$ is the closing price.

Based on $\hat{R}_{t+1}$ from a model trained with $L_j$, we rank stocks and select the top-$k$ for an equally weighted portfolio held during day $t+1$, with daily rebalancing~\cite{ci-sthpan}. This allows isolating the influence of $L_j$ on portfolio performance. 

\subsection{Model Architecture}
\label{sec:model_architecture}
The experiments utilize the PortfolioMASTER model architecture, inspired by \citet{master}, adapted for stock return forecasting. 
PortfolioMASTER employs alternating blocks of temporal self-attention (processing each stock's history independently) and spatial self-attention (modeling cross-stock relationships at each time step). The input features (daily return and turnover) for each stock over a lookback window $T$ are first projected to a model dimension $D$. Positional encoding is added before the input passes through spatio-temporal encoder layers. A final attention based temporal aggregation mechanism, characteristic of MASTER, produces a representation for each stock, which is then decoded to predict next day returns. Key hyperparameters, including model dimension, number of encoder layers, and attention heads, were selected based on performance across a predefined grid of values, tuned separately for each loss function.

\subsection{Investigated Loss Functions}
\label{sec:loss_functions}
We evaluate a diverse set of loss functions to train our model, categorized as pointwise, combined pointwise-pairwise, and listwise. Let $\mathbf{y} = [y_1, \dots, y_N]$ be the true next-day returns and $\hat{\mathbf{y}} = [\hat{y}_1, \dots, \hat{y}_N]$ the model's predictions. An overview of these functions is presented in Table~\ref{tab:loss_function_formulas_simplified}.

The baseline pointwise loss is Mean Squared Error (MSE):
\begin{equation}
    L_{\text{MSE}}(\hat{\mathbf{y}}, \mathbf{y}) = \frac{1}{N} \sum_{i=1}^{N} (\hat{y}_i - y_i)^2.
    \label{eq:mse_defined_here} 
\end{equation}

Combined losses integrate MSE with a pairwise component $L_{\text{PairwiseComponent}}$:
\begin{equation}
    L_{\text{Combined}} = (1-\lambda) \cdot L_{\text{MSE}} + \lambda \cdot L_{\text{PairwiseComponent}},
    \label{eq:combined_loss_defined_here} 
\end{equation}
where $\lambda \in [0,1]$ is a weighting hyperparameter. Pairwise components are typically summed over the set of valid pairs $P_{\text{valid}} = \{ (i,j) \mid i \neq j, y_i \neq y_j \}$ and then normalized by dividing by $|P_{\text{valid}}|$.
The specific pairwise components investigated include Hinge, Margin Ranking, Bayesian Personalized Ranking (BPR), and RankNet. Additionally, we explore two Weighted Hinge variants (WHR1/WHR2), which assign greater importance to top-ranked stocks by applying weights based on rank $w_k$ to the Hinge loss terms; WHR1 uses a linear weighting scheme based on rank, while WHR2 uses an exponential one.
Detailed formulas for these components are in Table~\ref{tab:loss_function_formulas_simplified}.

The listwise approach, ListNet, calculates cross-entropy between softmax-transformed true and predicted scores. First, for both the true scores $\mathbf{y}$ and predicted scores $\hat{\mathbf{y}}$, probability distributions are computed. For a generic score vector $\mathbf{x}$ (which will be either $\mathbf{y}$ or $\hat{\mathbf{y}}$), the probability of item $k$ is given by:
\begin{equation}
    P_{\mathbf{x}}(k; \tau) = \frac{\exp(x_k / \tau)}{\sum_{j=1}^{N} \exp(x_j / \tau)},
    \label{eq:listnet_prob_dist}
\end{equation}
where $\tau$ is a temperature parameter. The ListNet loss is then the cross-entropy between $P_{\mathbf{y}}(k; \tau)$ and $P_{\hat{\mathbf{y}}}(k; \tau)$:
\begin{equation}
    L_{\text{ListNet}}(\hat{\mathbf{y}}, \mathbf{y}; \tau) = - \sum_{k=1}^{N} P_{\mathbf{y}}(k; \tau) \log(P_{\hat{\mathbf{y}}}(k; \tau)).
    \label{eq:listnet_final_loss}
\end{equation}

An overview of all investigated functions and their specific pairwise components is presented in Table~\ref{tab:loss_function_formulas_simplified}.

\section{Experimental Setup}
\label{sec:experimental_setup}
\textbf{Dataset and Features.} We use daily stock market data for $N=110$ S$\&$P 500 stocks, selected as the top 10 by market capitalization from each of the 11 GICS sectors. The data spans from  2015-01-03 to 2024-12-03. The input features ($F=2$) for each stock are its daily return and daily turnover (volume/price), normalized per stock using scalers from the training set. We use a lookback window of $T=20$ trading days.

\textbf{Training and Evaluation.} The data is split chronologically: 70\% for training, 15\% for validation, and 15\% for testing.
Models are trained for up to 50 epochs using the AdamW optimizer with weight decay, employing early stopping based on validation loss and a learning rate scheduler. For each investigated loss function, hyperparameters (including model parameters like dropout, $d_{model}$, $d_{ff}$, learning rate, and loss specific parameters like $\lambda$, margin $m$, scale $\alpha$, or temperature $T$) were tuned via grid search, minimizing the respective loss on the validation set. 

\textbf{Portfolio Simulation and Metrics.} We simulate a daily rebalanced, long only, top-$k$ portfolio ($k=5$) with equal weighting. Performance is evaluated using:
\begin{itemize}
    \item \textbf{Portfolio Metrics:} Cumulative Return (CR), Annualized Return (AR), Annualized Volatility (AV), Annualized Sharpe Ratio (SR, with a risk-free rate of 4.3\%), and Maximum Drawdown (MDD).
    \item \textbf{Predictive Quality Metrics:} Mean daily cross sectional Spearman Information Coefficient (IC Spearman), its Information Ratio (ICIR Spearman) and Precision@5 (P@5, fraction of top-5 predicted stocks with positive actual returns). The test set loss (MSE) is also reported.
\end{itemize}

\section{Results and Discussion}
\label{sec:results_discussion}
\begin{table*}[htbp!]
\centering
\caption{Performance Metrics for PortfolioMASTER Trained with Different Loss Functions on the S$\&$P 500 Test Set (Top-5 Stocks).}
\label{tab:main_results}
\resizebox{\textwidth}{!}{%
\begin{tabular}{@{}lrrrrrrrrrrrrr@{}}
\toprule
Loss Function & CR(\%) & AR(\%) & AV(\%) & SR & MDD(\%) & IC(Sp.) & Std IC(Sp.) & ICIR(Sp.) & P@5 & TestLoss(MSE) \\
\midrule
MSE & 79.28 & 14.78 & 15.79 & 0.6637 & -19.58 & 0.0754 & 0.0994 & 8.8076 & 0.3576 & \textbf{0.00286} \\
Hinge & 82.90 & 15.33 & 15.79 & 0.6984 & -19.58 & 0.0762 & 0.0991 & 8.8432 & 0.3586 & 0.00301 \\
Margin & \textbf{89.07} & \textbf{16.23} & 15.85 & \textbf{0.7529} & -18.33 & 0.0758 & 0.0989 & 8.8520 & \textbf{0.3593} & 0.00632 \\
BPR & 85.68 & 15.74 & 15.89 & 0.7200 & \textbf{-15.77} & 0.0733 & 0.1005 & 8.6915 & 0.3578 & 0.01145 \\
RankNet & 80.78 & 15.01 & 15.82 & 0.6771 & -18.97 & \textbf{0.0767} & 0.0991 & \textbf{8.8422} & 0.3586 & 0.01909 \\
WHR1 & 82.40 & 15.25 & 15.78 & 0.6938 & -19.54 & 0.0763 & 0.0991 & 8.8448 & \textbf{0.3593} & 0.00352 \\
WHR2 & 81.84 & 15.17 & 15.83 & 0.6866 & -19.74 & 0.0764 & 0.0991 & 8.8470 & 0.3582 & 0.00300 \\
ListNet & 87.41 & 16.00 & \textbf{15.79} & 0.7407 & -18.36 & 0.0761 & \textbf{0.0990} & 8.8482 & \textbf{0.3595} & 1.01212 \\
\bottomrule
\end{tabular}%
}
\end{table*}

We evaluated the PortfolioMASTER model trained with eight different loss function configurations on the S\&P 500 test set. Table~\ref{tab:main_results} summarizes the portfolio performance and predictive quality metrics.

\textbf{Portfolio Performance Analysis.}
Our results show that the choice of loss function significantly impacts portfolio performance. The Margin loss achieved the highest Annualized Return (16.23\%) and Sharpe Ratio (0.7529). ListNet also performed strongly, with an AR of 16.00\% and SR of 0.7407, along with the lowest Annualized Volatility (15.79\%).
Notably, BPR yielded the lowest Maximum Drawdown (-15.77\%), suggesting that models trained with this objective offer better risk control, despite a slightly lower SR (0.7200). The baseline MSE performed reasonably, but was outperformed by several ranking oriented losses in terms of risk adjusted returns. The weighted hinge variants (WHR1, WHR2) and the standard Hinge loss offered incremental improvements over MSE but did not reach the overall portfolio profitability of Margin, ListNet, or BPR for this top-5 strategy.

\textbf{Predictive Quality vs. Portfolio Outcomes.}
An interesting finding is that the predictive quality metrics were quite consistent across most loss functions. For instance, IC Spearman (measure of ranking quality) hovered around 0.073-0.077, and P@5 was consistently near 0.358-0.359. RankNet achieved the highest IC Spearman (0.0767), yet its portfolio AR and SR were only moderate. Conversely, Margin and ListNet, which excelled in portfolio metrics, did not show markedly superior ICs scores. This suggests that while a certain level of predictive ranking ability is necessary, the specific design of the loss function plays a crucial role in translating these predictive signals into profitable portfolio decisions. This is particularly true regarding how the loss function penalizes different types of ranking errors or which parts of the rank distribution it emphasizes. The high test set loss (MSE) for ListNet is expected, as it does not directly optimize for the accuracy of return value predictions but rather for the quality of the entire ranked list.

\textbf{Impact of Loss Function Design.}
Pairwise losses that explicitly model preferences or margins between stocks (Margin, BPR) appear to be effective. The Margin loss, by enforcing a margin for correctly ranked pairs, might encourage the model to make more confident distinctions between stocks, leading to better top-k selections. ListNet, by considering the entire list, may capture more global ranking patterns that are beneficial for portfolio construction. The effectiveness of BPR in reducing draw downs could stem from its focus on correctly ordering preferred (better-performing) items, potentially leading to more stable top-k selections during volatile market periods.

\section{Conclusion}
\label{sec:conclusion}
Our research shows that the choice of specific loss function  used to train Transformer model for ranking stocks affects how well an investment portfolio performs. While standard MSE loss offers a good starting point, more advanced loss functions designed for ranking especially Margin, ListNet and BPR can achieve better returns when considering risk, and also offer improved protection against large losses. 

We observed that strong portfolio performance does not always directly correlate with the highest scores on standard predictive quality metrics. This highlights that the loss function's ability to shape decision making for top-k selection is paramount. 
We contribute a comprehensive benchmark that clearly demonstrates how different loss functions impact a Transformer model's ability to learn and translate predictive signals into profitable, rank based stock selection strategies. These findings offer practical guidance for both researchers and practitioners: carefully choosing and adjusting ranking-specific loss functions is a vital step in creating more effective quantitative trading strategies using Transformer models.

Future work could extend this analysis in several key directions. This includes investigating a wider range of listwise loss functions and developing methods for the automated tuning of weights in combined loss functions. Furthermore, the generalizability of our findings should be validated on a larger dataset, such as the full list of S$\&$P 500 stocks, and tested under different market conditions or with different rules for building portfolios.

\section{GenAI Usage Disclosure}
The authors acknowledge the use of generative AI tools (specifically ChatGPT and Gemini models) during the preparation of this manuscript for assistance with text generation, code adaptation, and language refinement. All generated content was reviewed and edited by the authors to ensure accuracy and alignment with the research presented.

\bibliographystyle{unsrtnat}
\bibliography{chapters/bibl}

\end{document}